\documentclass[review]{elsarticle}

\usepackage{lineno,hyperref}
\usepackage{amsthm}
\usepackage{array}
\usepackage{multirow}
\usepackage{enumerate}
\usepackage{amsmath,amssymb,amsfonts}
\usepackage{algorithmic}
\usepackage{graphicx}
\usepackage{textcomp}
\usepackage{xcolor}
\journal{Neurocomputing}

\begin{document}

\begin{frontmatter}

\title{Hyperbolic Node Embedding for Signed Networks}


\author[address1,address2]{Wenzhuo Song}
\ead{songwz17@mails.jlu.edu.cn}
\author[address3]{Hongxu Chen}
\ead{Hongxu.Chen@uts.edu.au}
\author[address1,address2]{Xueyan Liu}
\ead{xueyanl17@mails.jlu.edu.cn}
\author[address4]{Hongzhe Jiang}
\ead{jianghongzhe@njfu.edu.cn}
\author[address1,address2]{Shengsheng Wang\corref{correspondingauthor}}
\ead{wss@jlu.edu.cn}
\cortext[correspondingauthor]{The corresponding author.}



\address[address1]{College of computer science and technology, Jilin University, Changchun, 130012, China}
\address[address2]{Key Laboratory of Symbolic Computation and Knowledge Engineering of Ministry of Education, Jilin University, Changchun, 130012, China}
\address[address3]{Advanced Analytics Institute, School of Computer Science, Faculty of Engineering and IT, University of Technology Sydney, NSW, 2007 Australia}
\address[address4]{ College of Mechanical and Electronic Engineering, Nanjing Forestry University, Nanjing, 210037, China}

\begin{abstract}
Signed network embedding methods aim to learn vector representations of nodes in signed networks. However, existing algorithms only managed to embed networks into low-dimensional Euclidean spaces whereas many intrinsic features of signed networks are reported more suitable for non-Euclidean spaces. For instance, previous works did not consider the hierarchical structures of networks, which is widely witnessed in real-world networks. In this work, we answer an open question that whether the hyperbolic space is a better choice to accommodate signed networks and learn embeddings that can preserve the corresponding special characteristics. We also propose a non-Euclidean signed network embedding method based on structural balance theory and Riemannian optimization, which embeds signed networks into a Poincaré ball in a hyperbolic space. This space enables our approach to capture underlying hierarchy of nodes in signed networks because it can be seen as a continuous tree. We empirically compare our method against six Euclidean-based baselines in three tasks on seven real-world datasets, and the results show the effectiveness of our method.

\end{abstract}


\end{frontmatter}

\section{Introduction}

The rapid development of the World Wide Web has enabled millions of people around the world to communicate, collaborate and share content on the web. To analyze such complex and heterogeneous data, researchers often represented this ubiquitous networked data as networks, where nodes and links represent the entities and their relationships, respectively \cite{newman2010networks}. To facilitate machine learning-based network analysis, Network Representation Learning (NRL) is widely studied to automatically learn low-dimension vector representation of nodes (a.k.a. node embedding) while preserving the main structural properties of the original network \cite{zhang2018network,goyal2018graph}. A common assumption of NRL is that the proximities among the vectors can reflect the relationships among the corresponding nodes such as similarity, type and polarity. 

Recently, Signed Network Representation Learning (SNRL) methods have gained considerable attention because the polarity of the links, i.e., positive and negative relationship among entities in a complex networked system \cite{tang2014distrust} can be naturally modelled in a signed network. It is reported that link polarities information can improve the performance of traditional tasks \cite{Leskovec2010,5714203} and thus signed networks have a wide range of application scenarios such as support/opposite relationships in social networks, synergistic/antagonistic drugs in Healthcare, and symbiotic/competitive animals in Ecosystem.

\begin{figure}
	\includegraphics[width=0.9\textwidth]{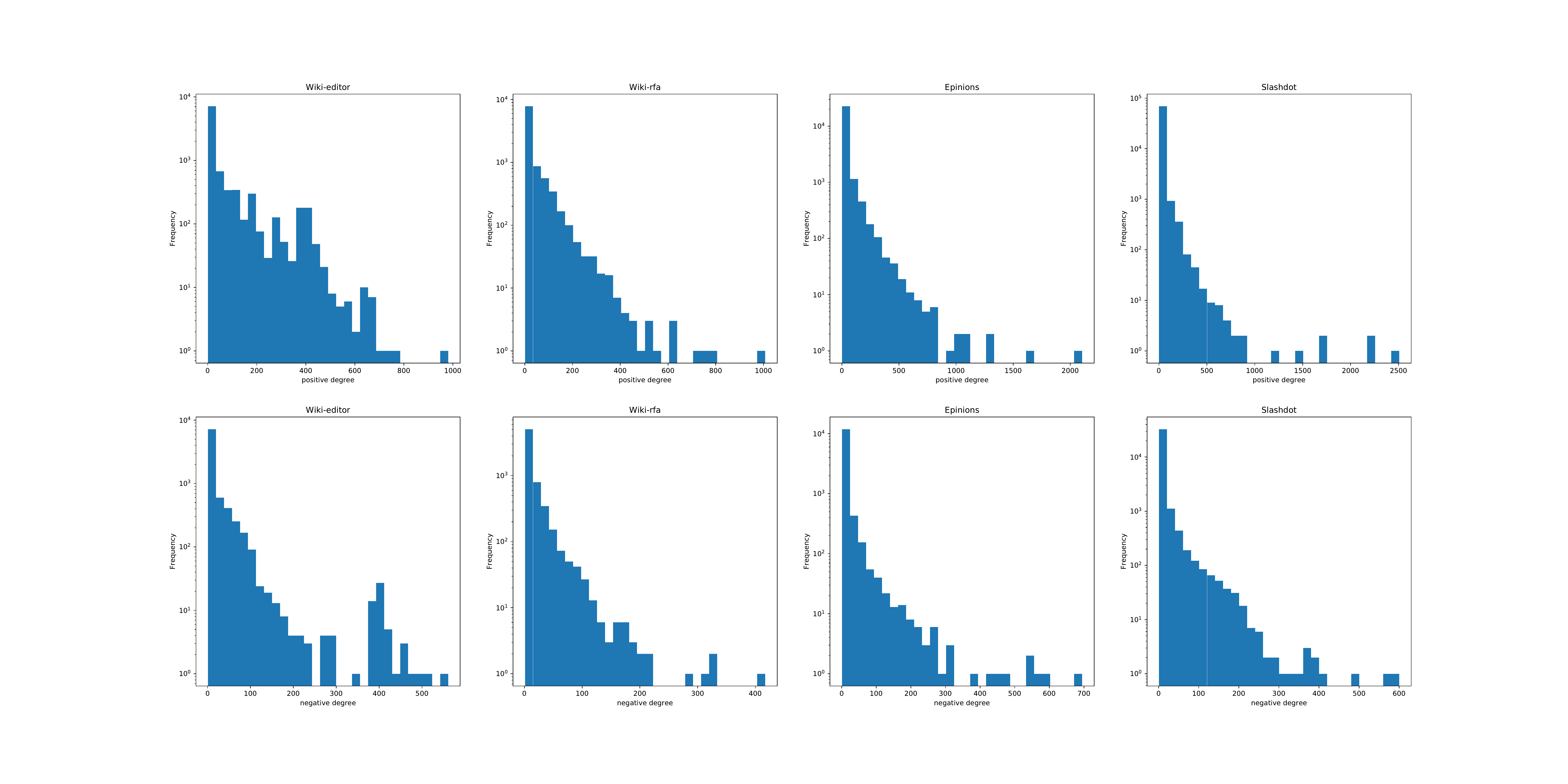}
	\caption{The degree distribution of positive and negative links of four real-world signed networks. All of them follow power-law degree distributions, which implies that these networks underlie hierarchical organizations.}\label{fig_powerlaw}
\end{figure}

Scale-free\footnote{https://en.wikipedia.org/wiki/Scale-free\_network} is an important property widely existed in real-world signed networks. In Figure \ref{fig_powerlaw}, we show that both positive and negative degree distributions of real-world signed networks in Table \ref{dataset} follow power-law distributions\footnote{We omit Epinions2 and Slashdot2 because they have similar results w.r.t reported versions.}. This result suggests that the hierarchy may be ubiquitous in many signed social networks because the scale-free property is the consequence of the underlying hierarchy \cite{clauset2008hierarchical,ravasz2003hierarchical}. For example, nodes with higher hierarchy are more likely to be connected by other nodes and vice versa.

However, none of the previous methods considers this intrinsic property of signed networks. Beseides, existing works learn node embeddings in Euclidean space where geometry constraints are imposed, and they may not be sufficient enough to model the data with latent hierarchies such as text, social networks and the web \cite{ganea2018hyperbolic, krioukov2010hyperbolic}. For example, considering the task of projecting a tree (can be seen as a simplified network) into a low-dimension Euclidean space where the distances between each pair of nodes are larger than a threshold. When the level of a tree increases, the dimensionality of the embedding space needs to be dramatically enlarged because it grows only polynomially while the size of the tree grows exponentially.

On the other hand, non-Euclidean NRL methods such as hyperbolic embedding methods have been proposed to model network structured data as well as unstructured data such as text \cite{ganea2018hyperbolic, tifrea2018poincar,de2018representation}. As a non-Euclidean space, hyperbolic space is suitable for modeling datasets with power-law distributions while capturing the latent hierarchical structures. One reason is that the hyperbolic space has a negative curvature so that space can expand exponentially for the radius\footnote{https://en.wikipedia.org/wiki/Hyperbolic\_space}. The hyperbolic space can be seen as a continuous tree such that the positions of nodes in the hyperbolic space can reflect the underlying hierarchical pattern of the networks (i.e., nodes closer to the center can serve as the root nodes of a network, while those nodes far apart the center are the leaf nodes).


Thus, in this work, we aim to answer the question: \textbf{Whether the hyperbolic space is a better choice to represent signed networks?} We propose a non-Euclidean representation learning method for signed networks named \textbf{H}yperbolic \textbf{S}igned \textbf{N}etwork \textbf{E}mbedding (\textbf{HSNE}). Specifically, we employ the structural balance theory from social theory to construct an effective objective function.  The key idea is that nodes connected by positive links are more similar than those connected by negative links. For example, in a politician social network, a positive link between two politicians represents they support each other, while a negative edge implies they are foes. The structural balance theory is consistent in many signed social network and can provide us with guidance to learn the node embeddings \cite{tang2016survey}. Secondly, we develop an efficient learning framework based on Riemannian stochastic gradient descent \cite{bonnabel2013stochastic}. The gradient calculation in hyperbolic space is more complex and time-consuming than that in Euclidean space. In this work, we sample a batch of nodes and their positive and negative neighbor nodes to train the model. We assume that these triples are independent of each other so that HSNE can scale to large scale dataset. 
Finally, we perform extensive experiments to evaluate the effectiveness of HSNE. We compare HSNE with Euclidean-based baselines on link sign prediction and reconstruction tasks, and the results show that our method can achieve similar or better performance in terms of the ability of generalization and capacity.

In summary, the main contributions of this work are:
\begin{enumerate}[1)]
	\item To the best of our knowledge, HSNE is the first SNRL method based on non-Euclidean geometry. It embeds the nodes of a signed network to a Poincaré ball in a hyperbolic space. We develop an efficient learning framework based on structural balance theory and Riemannian stochastic gradient descent.
	\item Benefited from the properties of hyperbolic geometry, our method can preserve the ubiquitous hierarchy and scale-free characteristics in signed networks, which are not considered in existing methods.
	\item To evaluate the effectiveness of HSNE, we perform experiments on three network analysis tasks on seven real-world datasets. The experimental results show that our approach performs similar or better than the state-of-the-art Euclidean counterparts.
	
\end{enumerate}

The remainder of this paper is organized as follows: In section 2, we propose a new non-Euclidean SNRL framework based on structural balance theory and hyperbolic embedding. We present experimental results on seven real-world networks in section 3, and in section 4 we briefly reviews the related works such as network representation learning and hyperbolic embedding. In the last section, we summarize the results of this work and discuss future works.

\section{The Proposed Method}

In this section, we discuss the proposed SNRL method named Hyperbolic Signed Network Embedding, which represents the nodes in a signed network with vectors in a Poincaré ball. Since hyperbolic space can be seen as a smooth tree, the locations of corresponding nodes can capture the underlying hierarchical structure of the network, and the distances between the vectors reflect their relationships such as proximity and polarity.

Let $G=(V,E)$ denote an undirected signed network, where $V=\{v_i...v_N\}$ and $E=\{e_{ij}\}_{i,j=1}^N$ are the sets of the $N$ nodes and $M$ edges in $G$. We can use a adjacency matrix $A^{N\times N}$ to represent $G$, where
\begin{equation}
\begin{split}
A_{ij}=
\begin{cases}
1,& \text{$e_{ij}$ is positive,}\\
-1,& \text{$e_{ij}$ is negative,}\\
0,& \text{otherwise.}
\end{cases}
\end{split}
\end{equation}
HSNE aims to learn a mapping function:
\begin{equation}
U=f(G)
\end{equation} 
where the $i$th item of $U=\{u_i\}_{i=1}^N$ is a vector in a $K$-dimension Poincaré ball $\mathbb{D}^K=\{u_i \in \mathbb{R}^K,||u_i||<1\}$ used to represent node $v_i$.   

To learn the node embeddings for signed networks, HSNE needs a) an objective function to measure the degree of fitting data, and b) an efficient learning framework used to map the nodes in signed networks to a Poincaré ball. 

\subsection{The Objective Function}

According to the objective function, SNRL methods can be roughly divided into similarity-based and social theory-based methods. Similarity-based methods first define the similarity between nodes based on the random walk or higher-order neighbor context and then use the distance between the node vectors to fit the similarity \cite{song2018learning,yuan2017sne,kim2018side}. These methods can consider higher-order dependency between nodes and often need a additional mapping function to obtain edge embeddings. Another approach usually employs the social theories, such as status theory and structural balance: ``the friend of my friend is my friend'' and ``the enemy of my enemy is my enemy'' \cite{zheng2015social}. Unlike similarity-based algorithms, structural balance usually treats negative links as the negation of positive, and the corresponding nodes should be far away \cite{wang2017signed,cygan2012sitting,kermarrec2011can}.

Recently, the structural balance theory is extended to ``one of my friend should be closer to me than my enemies'' \cite{cygan2012sitting}. This assumption relaxes the original definition and has successfully applied to embed the nodes in signed networks to a Euclidean space \cite{kermarrec2011can,wang2017signed}. Mathematically, given a node triple, i.e., three nodes $v_i$, $v_j$ and $v_k$ in $G$ where $A_{ij}=+1$ and $A_{ik}=-1$, the extended structural balance theory can be written as:
\begin{equation}\label{balance}
d(u_i,u_j) \leq d(u_i,u_k) - \lambda 
\end{equation}
where $d:(\mathbb{R}^K,\mathbb{R}^K)\rightarrow\mathbb{R^+}$ denotes the distance function between two vectors, and $\lambda>0$ is a hyperparameter, meaning how much closer a friend is than an enemy.

By punishing the triples that do not meet Equation \ref{balance}, we can get the objective function of HSNE:
\begin{equation}\label{eq:obj}
\mathop{\arg\min}_{U} \sum_{(v_i,v_j,v_k) \in T}\max(0,  d(u_i,u_j) -d(u_i,u_k) + \lambda)
\end{equation}
where $T = \{(v_i,v_j,v_k)|v_i,v_j,v_k \in V, A_{ij}>0, A_{ik}<0\}$ is the triple set sampled from $G$.

Next, by minimizing above objective function through gradient descent algorithms, we can get the node representations $U=\{u_i\}_{i=1}^N$. However, the optimization process has following two problems. First, the embeddings of some nodes in the network cannot be optimized since many nodes in signed networks have only positive links. For example, if $v_j$'s unique anchor node $v_i$ has only positive neighbor nodes, $v_j$ will not appear in T, so that its vector representation will not be optimized by HSNE. Second, in HSNE, $U$ and $d$ are defined in non-European space, so traditional gradient-based methods based on Euclidean space in previous works cannot be used to optimize Equation \ref{eq:obj}.

To solve the first problem, we relax the concepts of positive/negative neighbor nodes, i.e., a friend/enemy of $v_i$ is not necessarily directly connected to $v_i$:
\begin{equation}
\hat{T} = \{(v_i,v_j,v_k)|v_i,v_j,v_k \in V, \hat{A}_{ij}>0, \hat{A}_{ik}<0\}
\end{equation}
where $\hat{A}$ indicates the extended adjacency matrix. It contains the links in the original training set, and few inferred links used to complement $T$. We can construct $\hat{A}_{ij}$ through the following methods: 
\begin{itemize}
	\item \textbf{random sampling}: If $v_i$ has no positive/negative neighbors, we randomly select a node from the original network;
	\item \textbf{virtual node}: Suppose there is a node $v_0$, which is the enemy (or friend) of all other nodes \cite{wang2017signed};
	\item \textbf{social theory}: We utilizes structural balance theory or status theory to predict the relationship of unknown pairs of nodes \cite{zheng2015social,leskovec2010signed}.
\end{itemize}

For the second problem, we employ the Riemannian manifold gradient algorithm detailed in the next subsection.

\subsection{Optimization}

Different from existing SNRL methods, the parameters and the distance in HSNE are defined on a Riemannian manifold, which means that we cannot directly utilize the optimization approaches of previous methods. Instead, in this work we employ Riemannian stochastic gradient descent \cite{bonnabel2013stochastic} to minimize Equation \ref{eq:obj}. The optimization process of HSNE includes the following three steps:

1) Computing the stochastic Euclidean gradient of objective function: Specifically, for each triple $(v_i,v_j,v_k)\in\hat{T}$ in the training set, the stochastic Euclidean gradient for $v_j$ is defined as:
\begin{equation}
\bigtriangledown^{E}=\frac{\partial L(U)}{\partial d(u_i,u_j)}\frac{\partial d(u_i,u_j)}{\partial u_j}
\end{equation}  
where 
\begin{equation}
L(U)=\max(0,  d(u_i,u_j) -d(u_i,u_k) + \lambda)
\end{equation}
and
\begin{equation}
d(u_i,u_j)=\cosh^{-1}(1+\frac{2\|u_i-u_j\|^2}{(1-\|u_i\|^2)(1-\|u_j\|^2)})
\end{equation}
is the Poincaré distance between $u_i$ and $u_j$, and finally we get
\begin{equation}
\begin{split}
\bigtriangledown^{E} =
\begin{cases}
\frac{4}{\beta\sqrt{\gamma^2-1}}&(\frac{\left \| u_i \right \|^2-2\left \langle u_i,u_j \right \rangle +1}{\alpha}u_i-\frac{u_i}{\alpha}),\\& \text{if $\lambda> d(u_i,u_k)-d(u_i,u_j)$,}\\
0,& \text{otherwise,}
\end{cases}
\end{split}
\end{equation}
where $\alpha=1-\left \| u_j \right \|^2$, $\beta=1-\left \| u_i \right \|^2$ and $\gamma=1+\frac{2}{\alpha\beta}\left \| u_i-u_j \right \|^2$. In a similar way we can get the partial derivatives of $u_k$.

2) Deriving the Riemannian gradient from the Euclidean gradient: Since the hyperbolic space models are conformal to the Euclidean space, the Poincaré metric tensor $g_\theta^H$ satisfies the following formula:
\begin{equation}
g_\theta^H=\lambda_\theta^2g^E
\end{equation}
where $\theta$ is a point in $\mathbb{D}^K$, $\lambda_\theta=(\frac{2}{1-\left \| \theta \right \|^2})^2$ is the conformal factor and $g^E$ is the Euclidean metric tensor.

Finally, the Riemannian gradient $\bigtriangledown^{H}_\theta$ can be calculate by 
\begin{equation}
\bigtriangledown^{H}_\theta=\frac{(1-\left \| \theta \right \|^2)^2}{4}\bigtriangledown^{E}
\end{equation}

3) Applying Riemannian stochastic gradient descent (RSGD). To estimate the parameters in Equation \ref{eq:obj}, the update formula of HSNE is:
\begin{equation}\label{eq:update}
\theta_{t+1}=\mathfrak{R}_{\theta_{t}}(-\eta_{t} \bigtriangledown^{H}_{\theta_t})
\end{equation}
where $\eta_{t}$ is the learning rate at iteration $t$ and $\mathfrak{R}_{\theta_{t}}(s)=\theta_{t}+s$ is a retraction operation. We can also choose $\mathfrak{R}_{\theta_{t}}$ derived in \cite{ganea2018hyperbolic} as:
\begin{equation}
\begin{array}{l}{\mathfrak{R}_{\theta_{t}}(s)=} \\ {\frac{\lambda_{\theta_{i}}\left(\cosh \left(\lambda_{\theta_{i}}\|s\|\right)+\left\langle\theta_{i}, \frac{s}{\|s\|}\right) \sinh \left(\lambda_{\theta_{i}}\|s\|\right)\right)}{1+\left(\lambda_{\theta_{i}}-1\right) \cosh \left(\lambda_{\theta_{i}}\|s\|\right)+\lambda_{\theta_{i}}\left\langle\theta_{i}, \frac{s}{\|s\|}\right\rangle \sinh \left(\lambda_{\theta_{i}}\|s\|\right)} \theta_{i}} \\ {+\frac{\frac{1}{\|s\|} \sinh \left(\lambda_{\theta_{i}}\|s\|\right)}{1+\left(\lambda_{\theta_{i}}-1\right) \cosh \left(\lambda_{\theta_{i}}\|s\|\right)+\lambda_{\theta_{i}}\left\langle\theta_{i}, \frac{s}{\|s\|}\right.} \sinh \left(\lambda_{\theta_{i}}\|s\|\right)} ,\end{array}
\end{equation}
but in the experiments we find that the results of the two operations are close, and the former is less computationally intensive.

Finally, we employ a $\operatorname{proj}$ operator to avoid abnormal points
\begin{equation}
\operatorname{proj}(\theta)=\left\{\begin{array}{ll}{\theta /\|\theta\|-\varepsilon} & {\text { if }\|\theta\| \geq 1} \\ {\theta} & {\text { otherwise }}\end{array}\right.
\end{equation}
where $\varepsilon$ is a hyperparameter with small value.

Note that HSNE can easily scale to large dataset, because (1) we assume that the triples in Equation \ref{eq:obj} used to train the model are independent and (2) the proposed learning framework adopts a stochastic gradient optimization framework. At each step, our method randomly samples a batch of node triples to calculate the Riemannian gradient of the parameters, and then update the model with Equation 12. In the case of a large network, we can specify a small batch size to reduce the amount of calculation of gradient calculation, so that HSNE can scale to large networks.


\section{Experiments}
In this section, we use two groups of experiments to verify the effectiveness of HSNE. 
In the first group, we compare the performance of HSNE with Euclidean SNRL baselines in two real-world tasks, i.e., link sign prediction and reconstruction. Link sign prediction aims to test the generality ability of the SNRL methods. It is a widely used evaluation task and can predict the polarity of relationships in a complex system such as social networks, e-commerce, and the web. Link sign reconstruction is to predict the signs of known links from the results of the methods. We design this task to test the ability of capacity of the SNRL methods to extract and store information.

The second group of experiments is to evaluate whether the learned node vectors can reflect the latent hierarchical structure in signed networks. HSNE embeds the nodes in a real-world network to a 2-dimensional Poincaré disk. We refer node vectors near the center to root nodes of the network, and those far from the center as leaf nodes of the network.



\subsection{Datasets}
To evaluate the performance of the SNRL algorithms in real-world tasks, we conduct experiments on following seven real-world networks:
\begin{itemize}
	\item \textbf{Wiki-editor} \cite{yuan2017sne} is a collaborative social network where each node is an editor of Wikipedia\footnote{https://www.wikipedia.org/}. A positive link between two editors represents most pages they co-edited are from the same category, and vice versa.
	\item \textbf{Wiki-rfa}\footnote{https://snap.stanford.edu/data/wiki-RfA.html} \cite{west2014exploiting} was originally crawled for Person-to-Person sentiment analysis by SNAP\footnote{https://snap.stanford.edu/index.html}. Each node in this network represents a Wikipedia editor who wants to become an administrator and a positive/negative link from editor $i$ to $j$ means $i$ voted for/against $j$.
	\item \textbf{Epinions1}\footnote{http://www.trustlet.org/extended\_epinions.html} \cite{massa2008trustlet} is a consumer review site founded in 1999. The positive (negative) links represent trust (distrust) relationship between two users, i.e., the nodes in the network.
	\item \textbf{Epinions2}\footnote{https://snap.stanford.edu/data/soc-sign-epinions.html} \cite{leskovec2010signed} was crawled by Stanford SNAP group. Compared with Epinions1, this network is larger and more sparse.
	\item \textbf{Slashdot1}\footnote{https://snap.stanford.edu/data/soc-sign-Slashdot081106.html} \cite{leskovec2010signed} is a technology news website that allows the users to submit stories and article links. Any user can mark other people as a friend (positive link) or foe (negative link).
	\item \textbf{Slashdot2}\footnote{http://konect.uni-koblenz.de/networks/slashdot-zoo} \cite{kunegis2009slashdot} is another network crawled from Slashdot. The nodes and links of this network have the same meanings as Slashdot1.
	\item \textbf{Correlates of War (CoW)}\footnote{http://mrvar.fdv.uni-lj.si/pajek/SVG/CoW/} \cite{doreian2015structural,zhao2018network} is a international relations network. There are 137 nodes in this network, and each node represents a country. They are connected by 1152 links, where positive and negative links denote military alliances and disputes, respectively.
\end{itemize}

We remove the isolated nodes and small connected components for each data. The primary statistical information of these networks can be found in Table \ref{dataset}.

\begin{table*}
	\centering
	\caption{\label{dataset} The summary of seven real world signed networks used in the experiments.}
	\resizebox{\columnwidth}{!}{
	\begin{tabular}{cccccc}
		\hline
		Data & \#node & \#positive links & \#nagetive links & average degree & directional \\
		\hline
		Wiki-editor & 20,198& 268,420 & 78,798 & 34.4 & False\\
		Wiki-rfa & 11,259 & 132,751 & 38,811 & 30.5& True\\
		Epinions1 & 27,503 & 265,249 & 46,799 & 22.7& True\\
		Epinions2 & 119,130 & 586,223 & 118,349 & 11.8& True\\
		Slashdot1 & 77,350 & 353,595 & 114,959 & 12.1& True\\
		Slashdot2 & 79,120 & 350,005 & 117,864 & 11.8& True\\
		CoW & 137 & 915 & 237 & 16.8& False\\
		\hline
	\end{tabular}
	}
\end{table*}

\subsection{Baseline Methods}

In the first group of experiments, we compare the performance of following SNRL methods to evaluate the effectiveness of our method:

\begin{itemize}
	\item \textbf{SC} \cite{kunegis2010spectral} is a spectrum-based method. It transposes the eigenvector matrix corresponding to the $K$ smallest eigenvalues of the Laplacian matrix of the network as a low-dimensional representation of each node.
	\item \textbf{SNE} \cite{yuan2017sne} adopts a random walk approach to obtain the context of a node, i.e., the signs of links and the nodes along the path. The node embeddings are then calculated based on their similarity of node context by the Log-bilinear model\cite{mnih2013learning}.
	\item \textbf{SiNE} \cite{wang2017signed} uses a structural balanced-based objective function and employes multi-layer neural networks to measure the distances between nodes.
	\item \textbf{SLF} \cite{xu2019link} considers neural and none relationship between node pairs in addition to observed positive and negative links. It is designed for sign prediction task and networks of any sparsity.
	\item \textbf{SIDE} \cite{kim2018side} develops a link direction and sign aware random walk framework to preserve the information along with multi-step connections. In our experiments, all terms, e.g., signed proximity and bias term, are used to represent the nodes. 
	\item \textbf{BESIDE} \cite{chen2018bridge} combines the social balance and status theories in a joint neural network. The basic idea is that these two social-psychologic theories can complement each other.
	\item \textbf{HSNE} is the proposed SNRL method. It embeds each node to a hyperbolic space where the nodes are spontaneously organized hierarchically.
\end{itemize}

These methods can be roughly divided into similarity-based and social theory-based methods, according to the way of organizing nodes in vector space. SiNE and HSNE are social theory-based methods and assume that a node should be closer to its friends than foes. Thus, we directly use Euclidean distance and Poincaré distance to predict the signs of links for SiNE and HSNE, respectively. Mathematically, we use $s(i,j)$ to represent the score that $A_{ij}$ is a positive link, which is defined as:
\[ s(i,j)=-d(i,j) \]
where $ d(i,j)=u_iu_j^T $ for SiNE and $ d(i,j)=\cosh^{-1}(1+\frac{2\|u_i-u_j\|^2}{(1-\|u_i\|^2)(1-\|u_j\|^2)})$ for HSNE.

For other methods, such as SC and SNE, the distance between two points only represents their similarity. In other words, two nodes farther away do not mean that a negative link connects them. In order to predict the type of links, these methods first design several functions to map node vectors to edge vectors and then train a classifier to predict the type of links. Following the settings in previous works \cite{grover2016node2vec,song2018learning,yuan2017sne}, in this work, we test five mapping functions listed in Table \ref{tab:mapping}. We also employ logistic regression as the classifiers and use their predict confidence scores to evaluate the results.


\begin{table}
	\centering
	\caption{
		\label{tab:mapping} The functions for network embedding methods which map two node vectors, i.e., $u_i$ and $u_j$, to an edge vector, i.e., $B_{ij}$. All functions are element-wised and output a low-dimension vector.
	}
	\centering
	\begin{tabular}{cc}
		\hline
		Operator & Definition\\
		\hline
		hadamard & $B_{ij} = u_i \ast u_j$\\
		l1-weight & $B_{ij} = |u_i - u_j|$ \\
		l2-weight & $B_{ij} = |u_i - u_j|^2$ \\
		concate & $B_{ij} = u_i : u_j$\\
		average & $B_{ij} = \frac{1}{2}(u_i + u_j)$ \\
		\hline
	\end{tabular}
\end{table}

\subsection{Evaluation Metrics}
The link sign prediction and reconstruction are essentially binary classification tasks. Since the number of the two types of links in each network is unbalanced (see Table \ref{dataset}), we use Macro-F1, Micro-F1 and Area under the curve (AUC) score on the test set to evaluate the performance of each method. 

\textbf{Macro F1 and Micro F1:} Let $tp$, $fp$, $tn$ and $fn$ denote true positives, false positives, true negatives and false negatives, the precision and recall are defined as:
\[ precision = \frac{tp}{tp+fp} \]
\[ recall = \frac{tp}{tp+fn} \]
F1 score is defined as the harmonic average of the precision and recall:
\[ F1=2\cdot \frac{precision\cdot recall}{precision+recall} \]
Let $F1^+$ and $F1^-$ be the F1 scores for positive and negative links respectively. Macro-F1 and Micro F1 are defined as
$$Macro F1=\sum_{s \in \{+,-\}}(\frac{1}{2}\cdot F1^s)$$
and
$$Micro F1=\sum_{s \in \{+,-\}}(\frac{c^s}{M}\cdot F1^s)$$
where $c^s$ is the number of links with label $s$ and $n$ is the total number of links.

To calculate F1 scores, SNRL algorithms need a binary classifier to predict the type of each link in the test set. For SiNE and HSNE, we adopt the grid search algorithm on the validation set to obtain thresholds with best results and then predicts the type of links on the test set as:
\[ 
\begin{cases}
A_{ij} = 1,& \text{if $s(i, j)> threshold$}\\
A_{ij} = -1,& \text{if $s(i, j)< threshold$}
\end{cases}
\]
For other algorithms, we follow their setup to calculate edge embeddings via node embeddings and five operators in Table \ref{tab:mapping}, and then train a logistic regression with the edge embedding and corresponding sign as labels on the validation set. Finally, we use the classifier to predict the sign of links in the test set.

\textbf{AUC Score:} The value of $AUC$ depends on the rankings of positive and negative links \cite{hanley1982meaning}. For each pair of links in test set $\{(e_{ij},e_{mn})|e_{ij} \in E_+, e_{mn} \in E_-\} $, we first get
\begin{equation}
\begin{split}
score(e_{ij},e_{mn}) =
\begin{cases}
1,& \text{if $s(i, j)> s(m, n$)}\\
0.5,& \text{if $s(i, j)= s(m, n$)}\\
0,& \text{if $s(i, j)< s(m, n)$}
\end{cases}
\end{split}
\end{equation}
and the final score is obtained by averaging over all pair of links, i.e.,
\begin{equation}
AUC = \frac{1}{{|E_+|}\cdot {|E_-|}}\sum_{i,j}^{|E_+|}\sum_{m,n}^{|E_-|}score(e_{ij},e_{mn})
\end{equation}

\subsection{Parameters Settings}

In the experiments, we use the grid search algorithm to tune the hyperparameters of each method on the validation set. For SNE, we vary the sample size $ss$ and path length $pl$ in $ss\in\{1,2,3\}$ and $pl\in\{15,20,25\}$, respectively. For SiNE, we test real triplets $\delta$ and virtual triplets $\delta_0$ in $\{0.5,1.0\}$ . For HSNE, we test the three methods to construct $\hat{A}_{ij}$ and find the way to add a virtual node that performs better and efficient. In the visualization task, we use random sampling method because the results are clearest. We set $\lambda=\{0.1,6.0,1.0,1.0,1.0,1.0\}$ in link sign prediction task for Wiki-editor, Wiki-rfa, Epinions1, Epinions2, Slashdot1 and Slashdot2 respectively, and $\lambda=0.1$ for all datasets in link sign reconstruction task. The dimension of vectors $K$ in the methods are all set to 20. For other parameters, we use the default settings suggested by the authors in the papers or the source codes.

\subsection{Link Sign Prediction and Reconstruction}
This section contains two groups of experiments, i.e., link sign prediction and reconstruction. In the first group, we hide 20\% links from each network (10\% as the validation set and the other 10\% as the test set) and use the remaining links as the training set. The network representation learning methods learn the node embeddings from the training set and then predict the signs of hidden links. Better prediction results mean that the algorithm has better generalization ability. In the second group, we use the same experimental setup as the link sign prediction task except that the training set and the test set are both the entire network. Intuitively, a higher reconstruction score shows the corresponding model can extract and preserve more information from the data. We use the first five data sets because the small size of the CoW leads to unstable results after hiding links. We run each methods 5 times and report the average performance on Table \ref{tab:sign1} and Table \ref{tab:sign2}.
\begin{table}[]
\centering

	\caption{\label{tab:sign1} The average results of the SNRL methods in link sign prediction task. For SiNE and HSNE, we directly use the distance between two nodes. For other methods, we report the best result of five mapping functions, and the bold numbers represent the best results o the algorithms on the test set.}
\resizebox{\columnwidth}{!}{
\begin{tabular}{cccccccc}
\hline
Methods & Metrics & Wiki-editor           & Epinions1      & Epinions2      & Slashdot1      & Slashdot2 & Wiki-rfa     \\\hline
SC      & mac F1  & 0.567                  & 0.504          & 0.556          & 0.480          & 0.482      & 0.523      \\
        & mic F1  & 0.765                 & 0.616          & 0.741          & 0.750          & 0.745         & 0.647    \\
        & AUC     & 0.619                 & 0.576          & 0.612          & 0.527          & 0.515       & 0.619      \\\hline
SNE     & mac F1  & 0.538                 & 0.465          & 0.469          & 0.494          & 0.483         & 0.476    \\
        & mic F1  & 0.587                  & 0.557          & 0.552          & 0.531          & 0.514      & 0.519      \\
        & AUC     & 0.614                  & 0.553          & 0.558          & 0.538          & 0.526        & 0.529    \\\hline
SiNE    & mac F1  & 0.793                 & 0.619          & 0.641          & 0.627          & 0.622         & 0.564    \\
        & mic F1  & 0.850                  & 0.794          & 0.803          & 0.723          & 0.708      & 0.706      \\
        & AUC     & 0.890                & 0.709          & 0.723          & 0.694          & 0.677         & 0.587     \\\hline
SIDE    & mac F1  & 0.807                 & 0.719          & 0.703          & 0.691          & 0.684          & 0.628   \\
        & mic F1  & 0.845                  & 0.814          & 0.800          & 0.737          & 0.731      & 0.732      \\
        & AUC     & 0.918                & 0.865          & 0.838          & 0.791          & 0.775          & 0.722    \\\hline
SLF     & mac F1  & 0.817                 & 0.774          & 0.782          & \textbf{0.767} & \textbf{0.770}    & 0.730\\
        & mic F1  & 0.857                   & 0.860          & 0.865          & 0.808          & 0.811      & 0.782     \\
        & AUC     & 0.928                & \textbf{0.918} & 0.901          & \textbf{0.880} & \textbf{0.879}     & 0.862\\\hline
BESIDE  & mac F1  & 0.826        & \textbf{0.795} & \textbf{0.818} & 0.759          & 0.760         & \textbf{0.741}    \\
        & mic F1  & 0.880        & \textbf{0.904} & \textbf{0.908} & \textbf{0.827} & \textbf{0.825}    & \textbf{0.836}\\
        & AUC     & 0.910         & 0.909          & \textbf{0.918} & 0.871          & 0.869         & \textbf{0.863}   \\\hline
HSNE    & mac F1  & \textbf{0.907}          & 0.78           & 0.808          & 0.756          & 0.752         & 0.711  \\
        & mic F1  & \textbf{0.936}           & 0.887          & 0.892          & 0.815          & 0.808      & 0.789    \\
        & AUC     & \textbf{0.97}            & 0.891          & 0.914          & 0.858          & 0.854      & 0.82   \\\hline
\end{tabular}
}

\end{table}

\begin{table}[]
\centering
	\caption{\label{tab:sign2} The results of the SNRL methods in link sign reconstruction task. The experimental setup is the same as link sign prediction task except that the training set and the test set are both the entire dataset. This group of experiments aims to compare the capacity of each algorithm to extract and preserve information of the original networks.}
\resizebox{\columnwidth}{!}{
\begin{tabular}{cccccccc}
\hline
Methods & Metrics & Wiki-editor         & Epinions1      & Epinions2      & Slashdot1      & Slashdot2      & Wiki-rfa  \\\hline
SC      & mac F1  & 0.557                & 0.513          & 0.569          & 0.520          & 0.521          & 0.530    \\
        & mic F1  & 0.766                 & 0.645          & 0.723          & 0.767          & 0.632         & 0.650    \\
        & AUC     & 0.643                  & 0.570          & 0.626          & 0.578          & 0.561        & 0.619    \\\hline
SNE     & mac F1  & 0.921                 & 0.797          & 0.468          & 0.717          & 0.749           & 0.495  \\
        & mic F1  & 0.942                  & 0.875          & 0.533          & 0.767          & 0.791        & 0.542    \\
        & AUC     & 0.988                  & 0.946          & 0.556          & 0.846          & 0.878       & 0.554     \\\hline
SiNE    & mac F1  & 0.806                 & 0.665          & 0.647          & 0.630          & 0.647           & 0.570  \\
        & mic F1  & 0.862                 & 0.836          & 0.812          & 0.744          & 0.752      & 0.720       \\
        & AUC     & 0.908                  & 0.750          & 0.724          & 0.698          & 0.705        & 0.598    \\\hline
SIDE    & mac F1  & 0.820                & 0.691          & 0.714          & 0.717          & 0.726          & 0.662    \\
        & mic F1  & 0.857                & 0.787          & 0.793          & 0.762          & 0.768          & 0.722    \\
        & AUC     & 0.913                 & 0.848          & 0.872          & 0.837          & 0.840          & 0.771   \\\hline
SLF     & mac F1  & 0.813                 & 0.799          & 0.796          & 0.825          & 0.824           & 0.794  \\
        & mic F1  & 0.850                 & 0.881          & 0.864          & 0.860          & 0.860          & 0.835   \\
        & AUC     & 0.926                 & 0.934          & 0.926          & 0.932          & 0.933         & 0.923    \\\hline
BESIDE  & mac F1  & 0.857               & 0.848          & 0.863          & 0.863          & 0.858            & 0.822   \\
        & mic F1  & 0.899                 & 0.931          & 0.927          & \textbf{0.903} & \textbf{0.900}   & 0.883 \\
        & AUC     & 0.948                 & \textbf{0.962} & 0.966          & \textbf{0.959} & \textbf{0.954}    & 0.938\\\hline
HSNE    & mac F1  & \textbf{0.968} & \textbf{0.893} & \textbf{0.91}  & \textbf{0.865} & \textbf{0.862} & \textbf{0.864} \\
        & mic F1  & \textbf{0.977} & \textbf{0.946} & \textbf{0.95}  & 0.901          & 0.898       & \textbf{0.906}    \\
        & AUC     & \textbf{0.994}  & \textbf{0.962} & \textbf{0.968} & 0.943          & 0.939    & \textbf{0.948}  \\\hline   
\end{tabular}
}
\end{table}
 From Table \ref{tab:sign1}, we can see that: 1) Among the first three classic algorithms, SC performs better than SNE on all networks except for Macro F1 and AUC scores on Slashdot datasets. The average Macro F1, Micro F1 and AUC of SC are 0.519, 0.711 and 0.578, respectively. SiNE has achieved significant improvements over SC and SNE in the sign prediction task. SiNE has an average increase of 13\%, 5\% and 14\% in Macro F1, Micro F1 and AUC scores compared to SC, respectively. These results show the effectiveness of the structural balance theory: ``A friend of mine should be closer to me than one of my enemy''. Thus, given two nodes, a shorter distance means that they are more likely to be friends and vice versa. Another advantage of SiNE is that it uses a multi-layer perceptron to fit the data. Therefore, it can further enhance the performance of the algorithm by using the powerful nonlinear feature transformation of deep neural networks. 2) Three advanced Euclidean approaches, i.e., SIDE, SLF and BESIDE, have better F1 and AUC scores on all datasets than the classic baselines. BESIDE can achieve the best results on Wiki-rfa and Epinions datasets, which shows that supplementing the structural balance with the status theory is a promising approach of SNRL methods. The results of SLF are similar to but slightly lower than BESIDE. Note that in our experiments, we have removed all neural links, which are not considered by other methods and may lead to a decrease in SLF's performance. SIDE is not doing well compared to the other two advanced approaches, which can be due to the operation of hiding links in the network destroyed some important paths. 3) The proposed method, HSNE achieves the best performance on Wiki-editor and similar results with the best Euclidean SNRL methods on other datasets. For Wiki-editor, HSNE has an increase of 8\%, 5\% and 6\% than BESIDE in terms of Macro F1, Micro F1 and AUC, respectively. For other datasets, the average differences of the metrics between HSNE and the best Euclidean SNRL methods are no more than 2\%, 3\% and 1\%. However, SIDE, SLF and BESIDE are designed for directional signed networks, i.e., they additionally need link directions to train and to make predictions. For networks whose link directions are trivial, e.g., co-edit relationship in Wiki-editor, they may make error predictions. To summary, HSNE can perform similarly to the state-of-the-art Euclidean-based algorithm, which shows its superiority in generalization ability of modelling signed networks. We also note that our framework is orthogonal with all existing method and thus has the potential to further improve the performance by incorporating more advanced techniques. 

From Table \ref{tab:sign2} we can see that most of the conclusions of the previous group of experiments are still correct in link sign reconstruction task, except for the following aspects: 1) The results of SNE have increased significantly in 4 of 6 datasets, because hidden links can break many random walk paths, causing SNE to perform poorly in link sign prediction task, so SNE may not achieve good results on partially visible networks. 2) HSNE achieves the best results on all datasets, except for the Micro-F1 and AUC on Slashdot. These results show that the proposed method has the best capacity ability to extract and preserve information.

\subsection{Visualization of Hierarchical Structure in Signed Networks}
\begin{figure*}
	\includegraphics[width=\textwidth]{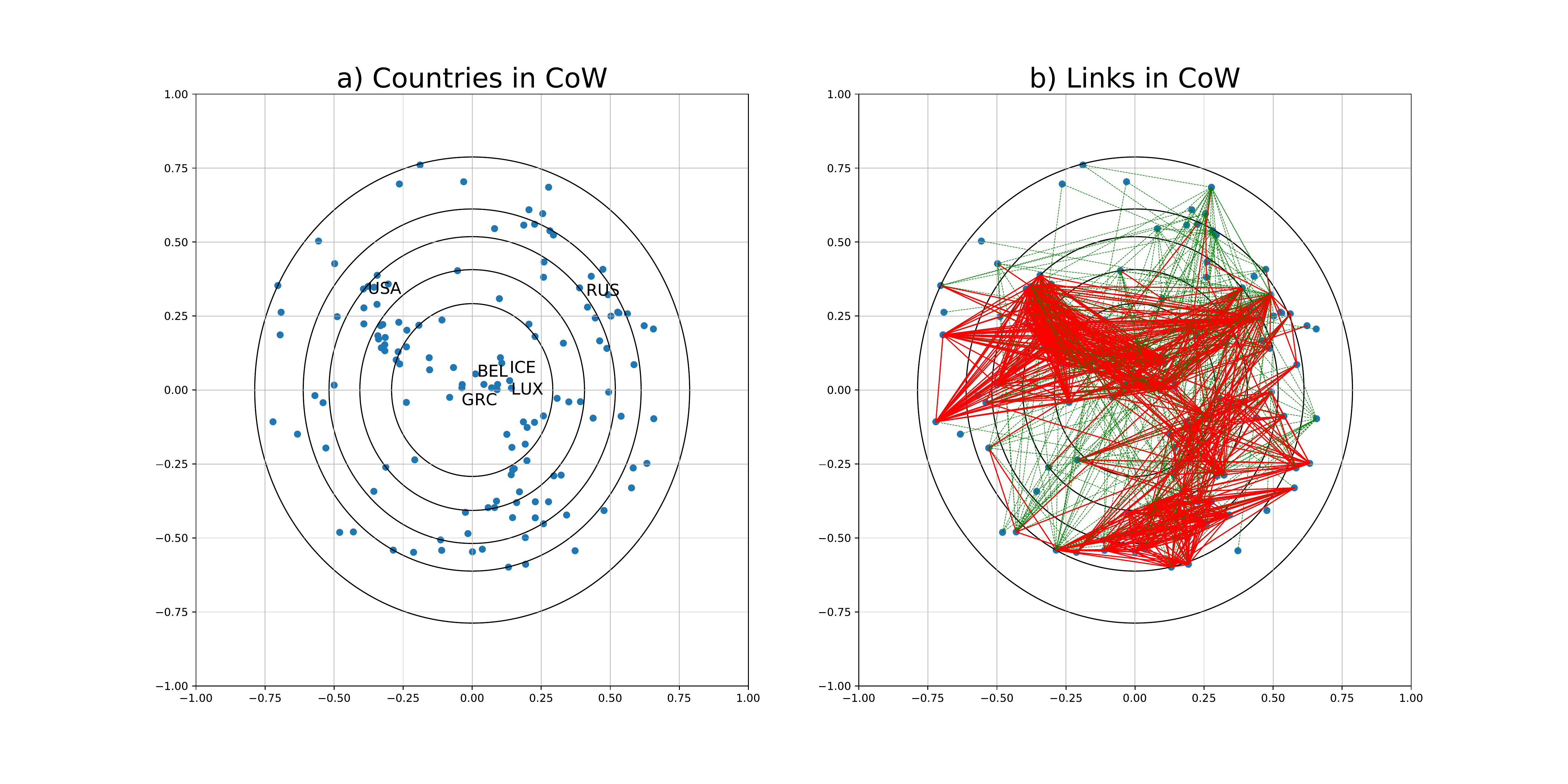}
	\caption{Visualization of countries in CoW network. Each point represents a node of the network, and the red and green lines represent positive or negative links between nodes, respectively. a) We divide the Poincaré disk into five parts by radius, with the same number of nodes in each. We can find that some neutral countries, such as Luxembourg, Iceland, are mapped inside the smallest circle. We refer these nodes as the root nodes of the network. b) The network generally consists of two communities where the positive links within the community are dense, i.e., the red areas in the upper left and lower right. The distance between nodes in different communities is large, meaning that they are more likely to be connected by negative links. On the other hand, all nodes are close to the root nodes.}\label{fig_emb}
\end{figure*}

\begin{figure}
    \centering
	\includegraphics[width=0.8\textwidth]{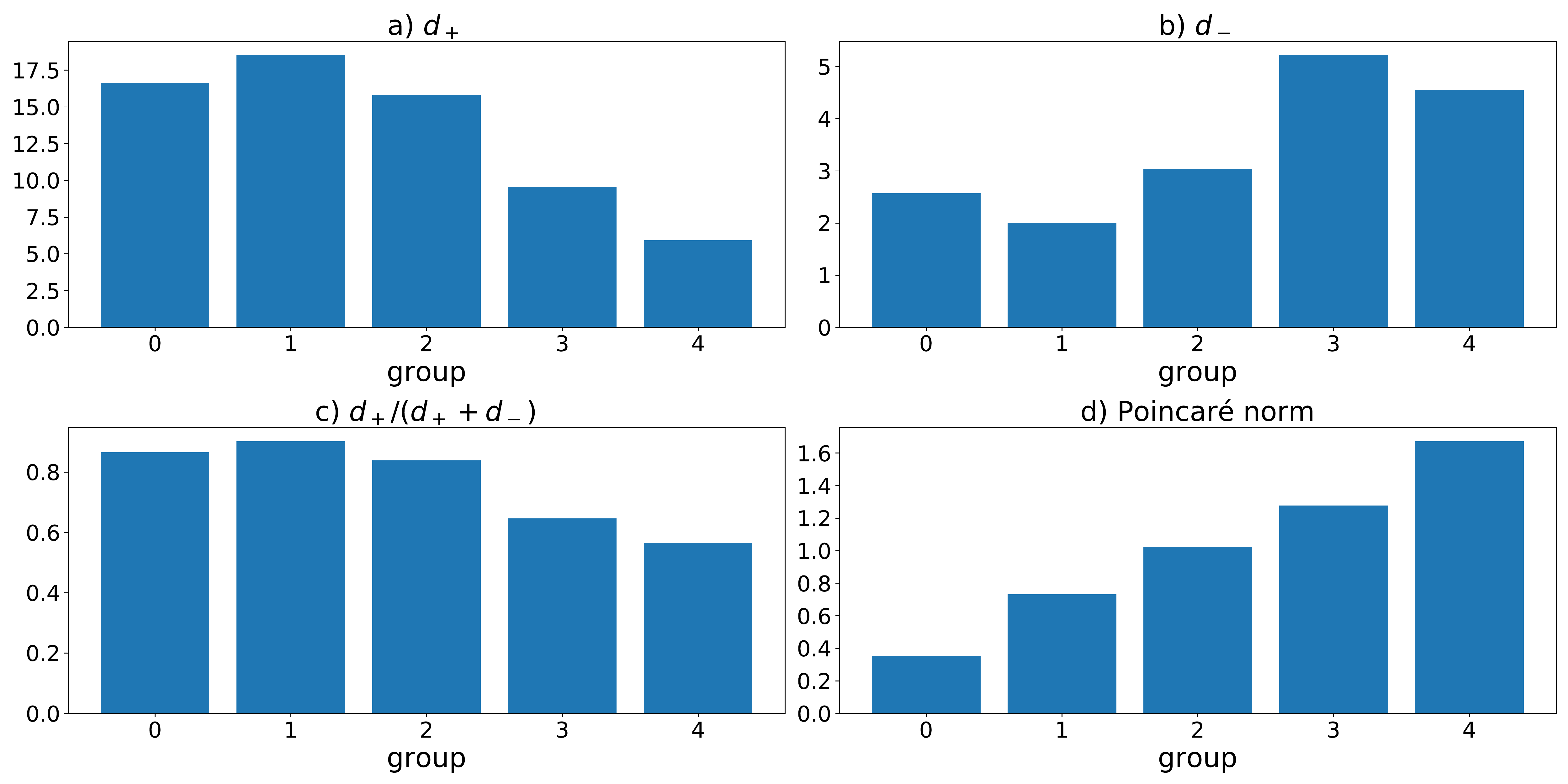}
	\caption{Comparison of basic information of the five groups of nodes in Figure \ref{fig_emb}, sorted by the average Poincaré norm: a) average positive degree $d_+$; b) average negative degree $d_-$; c) radio $d_+/d_-$, and d) average distance from the center, i.e., Poincaré norm}\label{fig_deg}
\end{figure}

In this group of experiments, we are interested in the ability of HSNE to capture the latent hierarchical structure in signed networks. We use HSNE to embed the nodes in CoW to a 2-dimension Poincaré disk. Since Poincaré disk can be seen as a continuous version of the tree, we refer the nodes close to the center of the disk as the root nodes of the network, and those nodes far from the center are leaf nodes. We divide the space into five areas according to the radius, with the same number of nodes in each area, and the results can be found in Figure \ref{fig_emb}. We also summarize some essential statistical characteristics of each group in Figure \ref{fig_deg}.

From Figure \ref{fig_emb} (a) and \ref{fig_deg}, we can find that some neutral countries, such as Luxembourg and Iceland, lie inside the group which is closest to the center. These countries have many friendly countries and few unfriendly countries, and can, therefore, be seen as a bridge or hub between many countries. As the distance from the center increases, the countries tend to form large alliances and are hostile to some countries. Some countries that are hostile to many countries are located at the margin of the space. From Figure \ref{fig_emb} (b) we can find that this network generally contains two communities, where the positive links within the community are dense, i.e., the two sectors at the upper left and lower right. This is because HSNE utilizes an objective function based on the structure balance theory, i.e., nodes connected by positive links are close to each other, and vice versa. If we consider these two groups as the two ``subtrees'' of the network and then regard the nodes near the center as the ``root'' of a tree, we can imagine the whole network as a binary tree. In this hierarchical structure, neutral countries have a higher level and serve as a bridge between countries. The countries with lower levels are not friendly to many countries except neutral countries. We can further recursively embed the nodes in each community, and finally, get the hierarchical structure of the network.

\begin{figure}
    \centering
	\includegraphics[width=0.5\textwidth]{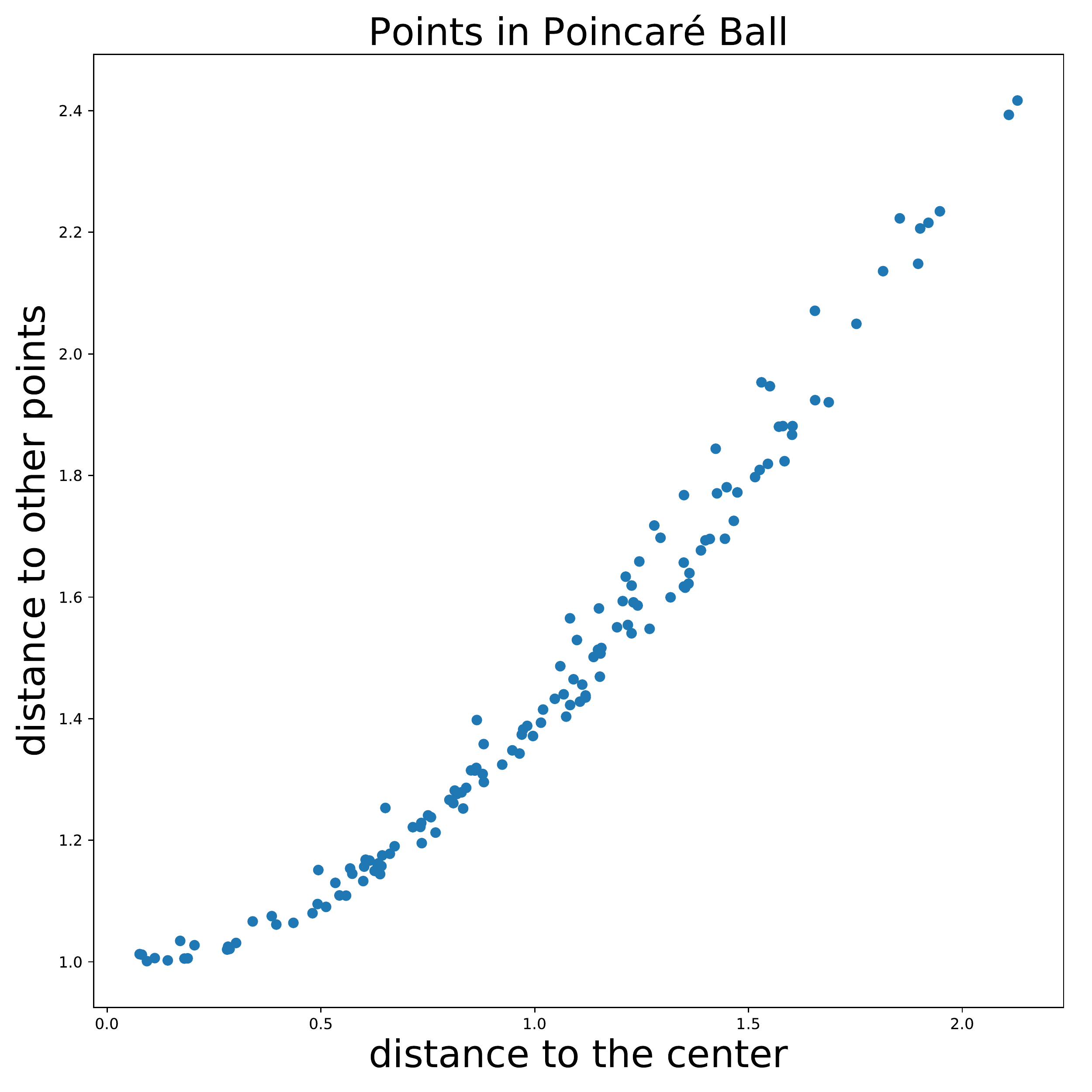}
	\caption{Visualization of the nodes in COW dataset. The x-axis represents the distance between a node and the center, and the y-axis is the average distance to other nodes. This figure illustrates how HSNE organizes nodes in a network in hyperbolic space.}\label{fig_dis}
\end{figure}

In the above experiments, we can conclude that HSNE can capture the underlying hierarchical structure in the signed networks. One may raise the question of why HSNE can represent the network hierarchically? To answer this question, we plot the nodes of CoW in Figure \ref{fig_dis} where the x-axis and y-axis represent the distance from the center and the average distance from other nodes, respectively. We can find that, in Poincaré ball, if a point is closer to other points, it is more likely to have a high level in the hierarchy, i.e., small Poincaré norm, and vice versa. Recall that the positive link of this network represents the friendship of two countries, while the negative link represents hostility. Thus, we can conclude that: 1) the countries near the center are more likely to be friendly with other countries because their average distances from other nodes are small. 2) As the number of unfriendly countries increases, the nodes gradually move away from the center. These countries also tend to form large alliances because nodes connected by positive links are more likely to be close to each other. 3) Nodes with many negative links are more likely to have lower levels in the hierarchy because they should be far away from many nodes in the network.

\section{Related works}
To analyze complex and non-linear systems \cite{sun2018adaptive,sun2019novel,sun2020fuzzy}, researchers often represented this ubiquitous networked data as networks, where nodes and links represent the entities and their relationships, respectively \cite{newman2010networks}. To facilitate machine learning-based network analysis algorithms, Network representation learning (NRL) methods aim to represent the nodes of a network as low-dimensional vectors \cite{zhang2018network,goyal2018graph}. The distances between the vectors reflect the relationship (such as similarity and weight) between the nodes so that they can be used to visualize the network \cite{maaten2008visualizing} and perform machine learning-based network analysis tasks such as node classification \cite{bhagat2011node}, link prediction \cite{liben2007link} and clustering \cite{ding2001min}. NRL methods can be divided into three categories: factorization methods, random walk-based techniques, and deep learning-based. Factorization methods such as GraRep \cite{cao2015grarep} and HOPE \cite{ou2016asymmetric} use dimensional reduction algorithms to process the matrix representation of the network, such as node adjacency matrix, node transition probability matrix, and Laplacian matrix. Random walk-based NRL methods such as DeepWalk \cite{perozzi2014deepwalk} and Node2vec \cite{grover2016node2vec} utilize random walk algorithms to obtain the context of each node and then embed them to low-dimension space by the similarity of contexts. This approach is especially useful when the network is large or local visible. Recently, deep learning-based methods have become popular due to its powerful ability to model high non-linear data \cite{wang2016structural,velivckovic2017graph,kipf2016variational}.

In recent years, signed network representation learning (SNRL) has gained considerable attention and proved effective in many tasks, such as node classification and sign prediction \cite{song2018learning}. Compared with NRL, SNRL needs to consider more complex semantic information such as the polarity of the links. Recent works point out that negative links have added values over positive links and can improve the performance of traditional tasks \cite{Leskovec2010,5714203}. In order to obtain effective node embeddings, SNRL methods have the following two basic steps: 1) designing an objective function to learn low dimension node embeddings. In this step, we can either interpret the negative links as the negation of positive or others of links \cite{tang2014distrust,ma2009learning,song2018learning}. 2) learning node embeddings using an efficient optimization framework. This step is essential to learn a mapping function from node proximity to low-dimensional Euclidean space. Popular approaches include word2vec \cite{mikolov2013efficient} and eigenvalue decomposition \cite{kunegis2010spectral}. 

Many recent works utilize advanced techniques to further improve the performance of learning framework, such as deep neural networks \cite{wang2017signed,song2018learning}, attention mechanism \cite{lu2019ssne}, graph convolutional operation \cite{derr2018signed} and negative sampling method \cite{islam2018signet}. For example, SiNE \cite{wang2017signed} and nSNE \cite{song2018learning} apply deep neural networks and metric learning to structural balance-based and node similarity-based objective functions, respectively. SIGNet \cite{islam2018signet} propose a novel social theory-based negative sampling technology to optimize classic similarity-based functions efficiently. However, most existing algorithms aim to map nodes to Euclidean space, which is an essential difference from our proposed algorithm.

Two general statistical characteristics are widely found in real-world networks: a) scale-free\footnote{https://en.wikipedia.org/wiki/Scale-free\_network} which refers to the degree distribution of a network follows a power-law distribution $p(k)=k^{-\gamma}$, where $2<\gamma<3$ typically, and b) a high degree of clustering which means many real-world networks are fundamentally modular. Ravasz et al. suggest that these two properties can be derived from the hierarchical organization of real-world networks \cite{ravasz2003hierarchical,albert2002statistical}. Recent works report that the hierarchical structure exists widely in real-world datasets \cite{adcock2013tree,traud2012social,adamic2005political}. 

Hyperbolic network embedding methods have attracted much attention because it is effective in modeling data with power-law distributions. These methods represent the nodes in a network with vectors in a hyperbolic space, where space expands exponentially with the radius since it has a constant negative curvature. Since hyperbolic space can be seen as a continuous tree, hyperbolic embedding can also reflect the underlying hierarchy of the data. Research on hyperbolic network embedding has just started in 2017. Nickel et al. \cite{nickel2017poincare} first introduce Poincaré ball model of hyperbolic space to learn node embeddings of social networks. Ganea et al. \cite{ganea2018hyperbolic} discuss the problem of embedding directed acyclic graphs (DAG) with a family of nested geodesically convex cones. Wang et al. \cite{HHNEAAAI19} map the nodes in heterogeneous information networks (HIN) to a Poincaré ball through meta-paths guided random walks. 


\section{Conclusion and Future Work}

In this paper, we develop a novel signed network embedding method based on hyperbolic space. This method automatically learns low-dimensional vector representations of nodes in a signed network to facilitate network analysis algorithms such as visualization, signed prediction and link reconstruction. We employ structural balance theory from social theory field to construct an objective function because many works have reported that most signed networks are balanced or tend to become balanced. This theory guarantees that similar nodes are mapped to close locations in embedding space, and dissimilar nodes are mapped to distant locations. Since the learning algorithms in previous SNRL methods cannot be applied to non-European space, we develop an efficient learning framework based on Riemannian stochastic gradient descent. This framework allows HSNE to scale to the large-scale dataset. We empirically use link sign prediction and reconstruction tasks to compare the performance of HSNE and Euclidean-based SNRL methods, and the results show that hyperbolic embedding can be a better space than Euclidean counterparts to represent signed networks. We also use HSNE to embed a real-world dataset CoW. We find our method places neutral countries near the center of the Poincaré disk. These countries have many friends and few enemies and therefore can be seen as the bridge or hub in the network. On the other hand, as the distance from the center increases, the countries tend to form alliances and are hostile to other countries. These results suggest that HSNE can extract a meaningful latent hierarchical structure from signed networks.

However, HSNE does not consider the rich node attribute information, which can further improve the performance intuitively. Besides, advanced technologies such as attention mechanism and graph convolutional operation are also not integrated into HSNE. They may provide a good boost in some domains or tasks. In the future, we will study the problem of recommendation systems based on hyperbolic embedding methods.

\section{Acknowledgements}

This work was funded by the Science \& Technology Development Project of Jilin Province, China [grant numbers 20190302117GX, 20180101334JC]; Innovation Capacity Construction Project of Jilin Province Development and Reform Commission [grant number 2019C053-3], China Scholarship Council under grant number 201906170208, 201906170205.

Declarations of interest: none.


\bibliography{mybibfile}

\end{document}